\documentclass[a4paper]{article}

\newcommand{\rem}[1]{}

\newcommand{\bre}{\begin{equation}}
\newcommand{\ere}{\end{equation}}

\newcommand{\ee}\]
\newcommand{\bra}{\begin{eqnarray}}
\newcommand{\era}{\end{eqnarray}}
\newcommand{\bfg}{\begin{figure}[hbtp]}
\newcommand{\efg}{\end{figure}}
\newcommand{\bver}{\begin{verbatim}}
\newcommand{\ever}{\end{verbatim}}
\newcommand{\bit}{\begin{itemize}}
\newcommand{\eit}{\end{itemize}}
\newcommand{\ben}{\begin{enumerate}}
\newcommand{\een}{\end{enumerate}}

\newcommand{\bGamma}{\boldsymbol\Gamma}

\newcommand{\btheta}{\boldsymbol\theta}

\newcommand{\given}{\: | \:}

\newcommand{\bo}{{\bf o}}

\newcommand{\bto}{\tilde {\bf o}}

\newcommand{\baa}{\begin{eqnarray*}}
\newcommand{\eaa}{\end{eqnarray*}}

\newcommand{\bs}{{\bf s}}

\newcommand{\cS}{{\cal S}}
\newcommand{\cT}{{\cal T}}

\newcommand{\hc}{\hat{c}}

\newcommand{\defined}{\triangleq}

\def\argmax{\mathop{\rm argmax}}

\def\defined{\: {\stackrel{\scriptscriptstyle \Delta}{=}} \: }

\newfont{\boldlarge}{msbm10 scaled 1100}

\newcommand{\comment}[1]{}

\newlength{\tmpbigbar}

\usepackage{spconf}

\usepackage{graphics}
\usepackage{amsthm}
\usepackage{amsmath}
\usepackage{amsfonts}
\usepackage{mathtools}
\usepackage{amssymb}
\usepackage{color}
\usepackage{stmaryrd}
\usepackage{cite}
\usepackage{color}
\usepackage{caption}
\usepackage{subcaption}
\usepackage{algorithm}
\usepackage[noend]{algpseudocode}

\usepackage{relsize}
\usepackage{array}
\usepackage{graphicx}
\usepackage{caption}
\usepackage{lmodern}
\usepackage{float}
\usepackage{nth}
\usepackage{multirow}
\usepackage{pgfplots}
\usepackage{hyperref}
\title{Simplified End-to-End MMI training and voting for ASR}
\name{Lior Fritz, David Burshtein}
\address{School of Electrical Engineering, Tel-Aviv University, Tel-Aviv 6997801, Israel \\
\href{mailto:lior.fritz@gmail.com}{lior.fritz@gmail.com}, \href{burstyn@eng.tau.ac.il}{burstyn@eng.tau.ac.il}}

\begin{document}

\maketitle

\begin{abstract}
A simplified speech recognition system that uses the maximum mutual information (MMI) criterion is considered. End-to-end training using gradient descent is suggested, similarly to the training of connectionist temporal classification (CTC). We use an MMI criterion with a simple language model in the training stage, and a standard HMM decoder. Our method compares favorably to CTC in terms of performance, robustness, decoding time, disk footprint and quality of alignments. The good alignments enable the use of a straightforward ensemble method, obtained by simply averaging the predictions of several neural network models, that were trained separately end-to-end. The ensemble method yields a considerable reduction in the word error rate.
\end{abstract}

\noindent\textbf{Index Terms}: neural networks, deep learning, hidden Markov models, connectionist temporal classification, speech recognition

\section{Introduction}
The hybrid approach of a hidden Markov model (HMM) and a deep neural network (DNN) \cite{hinton2012deep,vesely2013sequence} presents state-of-the-art results \cite{chan2015deep,xiong2016achieving} in automatic speech recognition (ASR). Yet the system is highly complicated. Connectionist temporal classification (CTC) \cite{graves2006connectionist,graves2014towards}, on the other hand, suggests a simple and scalable training procedure. It has been shown that with enough data, CTC can achieve state-of-the-art results \cite{hannun2014deep,amodei2016deep}. Recently, sequence-to-sequence models with attention \cite{chan2015listen,bahdanau2016end,chan2016latent,chorowski2016towards} have shown promising results, without the need for an external language model (LM) during decoding, minimizing the big disk footprint of the LM. Yet, attention models still fail to surpass phoneme-based CTC models, when an external LM is applied to both methods (compare \cite{chorowski2016towards} with \cite{wang2017residual}, also see our results). In addition, attention models impose long training times \cite{chan2016latent}. Moreover, since attention models are massively prone to overfitting, several heuristics take place in order to obtain good results \cite{chorowski2016towards,wu2016google,lamb2016professor}.
However, there are some issues with CTC. First, exact decoding is computationally intractable, and one needs to use some approximation \cite{graves2006connectionist}. To improve performance, some blank suppression method \cite{sak2015learning} or prior normalization \cite{miao2015eesen} is usually applied, which may be data-dependent \cite{kanda2016maximum}.
In addition, CTC does not excel in providing a good alignment between the input and output sequences, posing challenges in ensemble applications \cite{wang2017residual,sak2015acoustic}.
In \cite{povey2016purely}, lattice-free training using the maximum mutual information (MMI) criterion \cite{bahl1986maximum} is suggested. It is shown that lattice-free MMI training is a powerful technique. However, the training procedure still requires the use of HMM-GMM systems and decision trees, and it uses multiple-states context dependent senones instead of one-state characters or context independent phonemes that are typically used in CTC systems. Hence, the system is much more complicated compared to CTC, with larger disk footprint requirements.

In this paper, we suggest a simplified training method using the MMI criterion. Similarly to CTC, we use simple end-to-end training of a one-state context independent phonetic system. However, unlike CTC, our derivation is formulated using the MMI approach with a simple LM in the training stage. The resulting objective function is similar to \cite{collobert2016wav2letter} with some differences. First, we present our training method using the MMI criterion. Second, we allow the integration of a scalable LM into the objective function. Decoding can be performed using a standard HMM decoder. We use the weighted finite state transducer (WFST) approach \cite{mohri2002weighted} and discuss some implementation issues. Speech recognition results on the Wall Street Journal (WSJ) \cite{paul1992design} corpus compare favorably with CTC. Our method presents faster decoding times, and a reduction of 43\% of disk footprint of the decoder graph. Another advantage of the model is the fact that it provides a reliable alignment between the input and output sequences. This allows us to use a simple and effective voting method, obtained by simply averaging the outputs of several trained DNNs. We demonstrate a relative reduction of 23\% in word error rate (WER), compared to a single model.

\begin{figure}[hb]
	\centering
	\includegraphics[width=\linewidth]{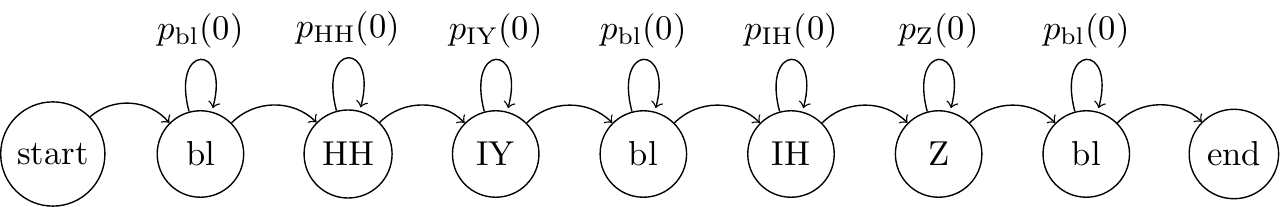}
	%    \captionsetup{justification=centering}
	\vspace{-15pt}
	\caption{HMM defined by the sentence ``he is''. bl denotes the blank state.}
	\label{fig:hi_all_hmm}
	\vspace{-20pt}
\end{figure}

\section{The Model}
Consider a phonetic hidden Markov model (HMM) for ASR \cite{rabiner1989tutorial}. Each word in a sentence is represented by a left to right HMM whose states are the phonemes. A blank state is inserted between words, and we reserve two states for sentence start and end.
We assume that if a word contains two consecutive identical phonemes, a blank state is inserted in-between.
Let $\{ \bo_t \}$ be the sequence of acoustic feature vectors representing some sentence. Define a time-extended feature vector, $\bto_t \defined \left( \bo_{t-F},\ldots,\bo_t,\ldots,\bo_{t+F} \right)$, for some integer $F$. We assume a left to right HMM  for the sequence $\{ \bto_t \}$. The underlying hidden state sequence  is $\{ s_t \}$, where $s_t \in \{ 0,1,\ldots,L-1\}$ and $L$ is the total number of possible states. The joint probability of $\bto \defined \{ \bto_t \}_{t=1}^{T}$ and $\bs \defined \{ s_t \}_{t=0}^{T}$ is given by
\begin{equation}
\label{eq:p_bto_bs}
P(\bto,\bs) = \prod_{t=1}^{T} p_{s_{t-1},s_t} \prod_{t=1}^{T} P(\bto_t \given s_t)
\end{equation}
where $p_{s_{t-1},s_t}$ is the transition probability between $s_{t-1}$ and $s_t$, and $P(\bto_t \given s_t)$ is the output probability. 

The transcript, $\bGamma$, of the sentence is the concatenated sequence of word states. As an example, consider the sentence ``he is''. Then the transcript is $\bGamma = \text{\{start, blank, HH, IY, blank, IH, Z, blank, end\}}$.
The sentence transcript, $\bGamma$, defines the left to right HMM, associated with the sentence.
%transitions are only allowed between neighboring states, or from a state to itself (self transition). That is,
%\begin{equation}
%P\left( \Gamma_k \given \Gamma_j \right) =
%\left\{
%\begin{array}{ll}
%p_{\Gamma_j}(0)              & \mbox{if $k = j$}\\
%p_{\Gamma_j}(1)              & \mbox{if $k = j + 1 $}\\
%0              & \mbox{otherwise}
%\end{array}
%\right.
%\end{equation}
%where $\Gamma_l$ is the $l$-th state in the transcript,
For each possible state $c=0,1,\ldots,L-1$ in our phonetic model, $p_c(0)$ denotes its self transition probability, and $p_c(1)$ denotes the probability of transition to the next state in the transcript, so that $p_c(0) + p_c(1) = 1$.
Figure \ref{fig:hi_all_hmm} describes the HMM defined by the sentence ``he is''.

Given the sequence of HMM states associated with some sentence, $\bs = \{ s_t \}_{t=0}^{T}$, the corresponding sentence transcript, $\bGamma$, can be obtained from $\bs$ by erasing repeated states as in CTC~\cite{miao2015eesen}.

To formulate an MMI criterion we need to define a probabilistic LM for $\bs$. This model also defines the distribution of the transcript random variable, $\bGamma$. As an optimal solution, one would use the ground-truth LM used in the decoding stage (or some degenerated form of it) \cite{vesely2013sequence,povey2005discriminative}, yet this would be computationally infeasible. As an alternative solution, we use a degenerated model of states $n$-grams, as in \cite{povey2016purely}.

% Given that at time $t-1$ the process is at state $c$, we remain at $c$ with probability $p_c(0)$, and make a transition to any other state $\hc$ with probability $p_c(1)q(c,\hc)$. That is,
Given that at time $t-1$ the process is at state $c$, we remain at $c$ with probability $p_c(0)$, and make a transition to any other different state $\hc$ with probability $ p_c(1) $ multiplied by the transition probability of the n-grams states model. In a bi-grams model, we define this probability as $ q(c,\hc) $. Later we show that a bi-grams model is already very powerful. For brevity, we describe our training method for a bi-grams model. That is,
\begin{align}
p_{c,\hc} = &
P\left( S_t=\hc \given S_{t-1}=c \right) \nonumber \\
= &
\left\{
\begin{array}{ll}
p_c(0)              & \mbox{if $\hc=c$}\\
p_c(1) \cdot q(c,\hc)    & \mbox{otherwise}
\end{array}
\right.
\label{eq:P_st_given_st1}
\end{align}
Note that due to the assumption that there are no consecutive phonemes, $ q(c, c) = 0 $. 

Now suppose also that some neural network (NN) recognizer produces the values $y_{t,s_t} \equiv y_{t,s_t,\btheta}$, which are estimates of \footnote{The base of all the logarithms in this paper is $e$} $\log P(s_t \given \bto_t)$, for $t=1,2,\ldots,T$. Here $\btheta$ are the parameters of the NN.
%Later on we consider a bidirectional recurrent NN (RNN).
Note that $\sum_{l=0}^{L-1} e^{y_{t,l}} = 1$ for all $t$.
By Bayes' law, \eqref{eq:p_bto_bs} can be rewritten as,
\begin{equation}
%\begin{align}
\label{eq:p_bto_bs_A}
P(\bto,\bs) 
%=&
%A \prod_{t=1}^{T} P(s_t \given s_{t-1}) \prod_{t=1}^{T} \frac{P(s_t \given \bto_t)}{P(s_t)}\\
%=&
=
A \left( \prod_{t=1}^{T} p_{s_{t-1},s_t} \right) \cdot
\exp\left\{
\sum_{t=1}^{T} \overline y_{t,s_t}
\right\} 
%\nonumber
%\end{align}
\end{equation}
where
$ \overline y_{t,s_t} \defined y_{t,s_t} - \omega_{s_t} $,
%$p_{s_{t-1},s_t} \defined P(s_t \given s_{t-1})$ is the transition probability from $s_{t-1}$ to $s_t$, 
$\omega_l \defined \log P(l)$ is the logarithm of the prior probability of the state $l$,
and
$ A = \prod_{t=1}^T P(\bto_t) $.

%To apply \eqref{eq:map_criterion}, we need to compute $P(\bto)$ and $P(\bto, \bGamma)$. By~\eqref{eq:p_bto_bs_A} and since, as explained, we may set $A=1$, the first term, 
By summing over all possible state sequences, we obtain 
\begin{equation}
P(\bto) 
=
A \sum_{\bs}
\left( \prod_{t=1}^{T} p_{s_{t-1},s_t} \right) \cdot
{\rm exp}\left\{ \sum_{t=1}^{T} \overline y_{t,s_t}\right\}
\end{equation}
%The transition probabilities are given by~\eqref{eq:P_st_given_st1}.

To compute $P(\bto,\bGamma)$ we use,
\begin{equation}
\label{eq:p_bto_and_Gamma}
P(\bto,\bGamma) 
=
A \sum_{\mathclap{\bs\in\cS(\bGamma)}}
\left( \prod_{t=1}^{T} p_{s_{t-1},s_t} \right)
\cdot
{\rm exp}\left\{ \sum_{t=1}^{T} \overline y_{t,s_t}\right\}
\end{equation}
where $\cS(\bGamma)$ are all state sequences $(s_0,s_1,\ldots,s_T)$ which are consistent with the given transcript, $\bGamma$.
%, and where we use the transition probabilities in~\eqref{eq:P_st_given_st1}.

\subsection{Training Procedure}
We train our model using a database of $N$ pairs $\left( \bGamma^n, \bto^n \right)$, $n=1,2,\ldots,N$, where $\bto^n$ is the $n$-th observation sequence, and $\bGamma^n$ is its associated transcript.
First, the values of the transition probabilities $ q(c,\hc) $ can be estimated from a training text using
\begin{equation}
	q(c,\hc) = N_{c,\hc} / N_c
\end{equation}
where $ N_{c,\hc} $  ($N_c$, respectively) is the number of occurrences of consecutive pairs of states $ c, \hc $ (states, $c$) in the text.
%Smoothing methods can also be incorporated.

To train the remaining model parameters we use the MMI criterion. The goal is to maximize the log-likelihood of the transcript given its observation sequence. That is, the MMI estimate, $\Psi_{\rm MMI}$, to the parameter vector, $\Psi = \left\{ \btheta, \left\{ p_j(0), p_j(1), \omega_j \right\}_{j\in\{0,1,\ldots,L-1\}}
\right\}$, is given by
\begin{equation}
\Psi_{\rm MMI} = \argmax_{\Psi} \sum_{n=1}^{N}
\log P_\Psi \left( \bGamma^n \given \bto^n \right)
\label{eq:map_criterion}
\end{equation}
%where
\begin{equation}
\log P \left( \bGamma \given \bto \right) = 
\log P \left( \bto, \bGamma \right) -
\log P \left( \bto \right) 
\label{eq:p_gamma_given_o}
\end{equation}
Since the term $A$ cancels out in \eqref{eq:p_gamma_given_o}, in the sequel we set $A=1$ without loss of generality.

Once the model has been trained, Decoding is preformed with a standard LM instead of the simple states $n$-gram LM used for training.

To compute $P(\bto,\bGamma)$ and its derivatives efficiently, first denote 
\begin{alignat}{2}
\alpha_t(k) &\defined
P \left(\cT(S_0,\ldots,S_t) \right.&& \in (\Gamma_0,\ldots,\Gamma_k)\\
& &&,\left.\bto_1,\ldots,\bto_t \right)\nonumber\\
\beta_t(k) &\defined 
P \left( \cT(S_t,\ldots,S_T)\right.&& \in (\Gamma_k,\ldots,\Gamma_K)\\
& &&,\left.\bto_{t+1},\ldots,\bto_T \given S_t = \Gamma_k\right)\nonumber
\end{alignat}
where $\cT(S_0,\ldots,S_t)$ is the transcript of the state sequence up to time $t$ and where $\Gamma_0,\ldots,\Gamma_k$ is the $k+1$ prefix of $\bGamma$. $\cT(S_t,\ldots,S_T)$ is the transcript of the state sequence from $ t $ to $ T $ and $\Gamma_k,\ldots,\Gamma_K$ is the $K-k+1$ suffix of $\bGamma$. Then $ P(\bto,\bGamma) = \sum\limits_{k=0}^K \alpha_t(k)\beta_t(k) $ for any $t$.

Now, $\alpha_t(k)$ and $ \beta_t(k) $ can be computed efficiently using the standard forward and backward time recursions,
\begin{align}
\alpha_t(k) = &
\left(
\alpha_{t-1}(k) p_{\Gamma_k}(0)\right. \label{eq:alpha_recur}\\
& \left. + \alpha_{t-1}(k-1) p_{\Gamma_{k-1}}(1) q(\Gamma_{k-1}, \Gamma_k)
 \right) e^{\overline y_{t,\Gamma_k}}\nonumber\\
\beta_t(k) = &
\beta_{t+1}(k) p_{\Gamma_k}(0) e^{\overline y_{t+1,\Gamma_k}} \label{eq:beta_recur}\\
& + \beta_{t+1}(k+1) p_{\Gamma_k}(1) q(\Gamma_k, \Gamma_{k+1}) \nonumber
e^{\overline y_{t+1,\Gamma_{k+1}}}
\end{align}

Our training algorithm is a stochastic gradient descent (SGD) that operates on mini-batches. To compute the gradient with respect to any element, $\theta_i$, of the parameter vector, $\btheta$, we use
\begin{equation}
\frac{\partial P(\bto, \bGamma)}{\partial \theta_i} =
\sum_{t,k} \frac{\partial P(\bto, \bGamma)}{\partial y_{t,k}}
\frac{\partial y_{t,k}}{\partial \theta_{i}}
\end{equation} 

We calculate $\frac{\partial y_{t,k}}{\partial \theta_{i}}$ using the back-propagation algorithm.
In addition, by \eqref{eq:p_bto_and_Gamma},
\begin{align}
\label{eq:dpog_ytr}
\frac{\partial P(\bto, \bGamma)}{\partial y_{t,r}} 
& = \sum_{k \: : \: \Gamma_k=r} \alpha_t(k) \beta_t(k)
% \\
% \label{eq:dpog_or}
% \frac{\partial P(\bto, \bGamma)}{\partial \omega_r} 
% &= -\sum_{t=1}^{T}\sum_{k \: : \: \Gamma_k=r} \alpha_t(k) \beta_t(k)\\
% \frac{\partial P(\bto, \bGamma)}{\partial p_r(0)} 
% &= \sum_{t=1}^{T}\sum_{k \: : \: \Gamma_k=r} \alpha_{t-1}(k) \beta_t(k) \exp\left\{
% \overline y_{t,\Gamma_k}\right\} 
% \\
% \label{eq:dpog_pr1}
% \frac{\partial P(\bto, \bGamma)}{\partial p_r(1)}
% &= \sum\limits_{t=1}^{T}\sum\limits_{k \: : \: \Gamma_k=r} \alpha_{t-1}(k) q(\Gamma_k, \Gamma_{k+1}) \\
% & \qquad\qquad\qquad\cdot \beta_t(k+1)\exp\left\{
% \overline y_{t,\Gamma_{k+1}} \right\} \nonumber
\end{align}
The derivatives of the other parameters in $ \Psi $ can be easily obtained similarly. Note that the values $e^{y_{t,l}}$ are constrained to be valid probabilities. To satisfy this constraint we use the standard approach of a softmax layer at the output of the NN. We use a similar approach on $p_j(0)$, $p_j(1)$ and $\omega_j$, in order to constrain them to be valid probablities.

To compute $P(\bto)$ and its derivatives efficiently, first denote
\begin{equation}
\tilde \alpha_t(c) \defined P \left(\bto_1,\bto_2,\ldots,\bto_t,S_t=c \right)
\end{equation}
\begin{equation}
\tilde \beta_t(c) \defined P\left(\bto_{t+1},\bto_{t+2},\ldots,\bto_{T} \given S_t=c\right)
\end{equation}
Then $ P(\bto) = \sum\limits_{c} \tilde\alpha_t(c)\tilde\beta_t(c) $ for any $t$. 
Now, $\tilde \alpha_t(c)$ and $\tilde \beta_t(c) $ can be computed efficiently
using standard forward and backward time recursions similarly to \eqref{eq:alpha_recur}-\eqref{eq:beta_recur}.
%sing,
%\begin{align}
%& \Biggl. + \sum\limits_{\hc \ne c} \tilde \alpha_{t-1}(\hc) p_{\hc}(1) q(\hc,c) \Biggr) %e^{\overline y_{t,c}} \nonumber \\
%%& + \sum\limits_{\hc \ne c} \tilde \beta_{t+1}(\hc) p_{c}(1) q(c,\hc) e^{\overline %y_{t+1,\hc}} %\nonumber
%\tilde \alpha_t(c) = & \Bigl( \Bigr.
%\tilde\alpha_{t-1}(c) p_{c}(0)\\
%& \Bigl. + \sum\limits_{\hc \ne c} \tilde \alpha_{t-1}(\hc) p_{\hc}(1) q(\hc,c) \Bigr) e^{\overline %y_{t,c}} \nonumber \\
%\tilde \beta_t(c) = & 
%\tilde\beta_{t+1}(c) p_{c}(0) e^{\overline y_{t+1,c}} \\
%& + \sum\limits_{\hc \ne c} \tilde \beta_{t+1}(\hc) p_{c}(1) q(c,\hc) e^{\overline y_{t+1,\hc}} %\nonumber
%\end{align}
Also,
\begin{align}
\frac{\partial P(\bto)}{\partial y_{t,r}} 
&= \tilde\alpha_t(r)\tilde\beta_t(r) \label{eq:dPOdy}
% \\
% \frac{\partial P(\bto)}{\partial \omega_r} 
% &=
% -\sum\limits_{t=1}^{T} \tilde\alpha_t(r)\tilde\beta_t(r) \label{eq:dPOdomega}\\
% \frac{\partial P(\bto)}{\partial p_r(0)} 
% &=  \sum\limits_{t=1}^{T}\tilde\alpha_{t-1}(r) \tilde\beta_t(r)\exp\left\{
% \overline y_{t,r} \right  \} \label{eq:dPOdpr0}\\
% \frac{\partial P(\bto)}{\partial p_r(1)}
% &= \sum\limits_{t=1}^{T}\sum\limits_{c \ne r} \tilde\alpha_{t-1}(r) q(r, c) \tilde\beta_t(c)\exp\left\{
% \overline y_{t,c}\right\} \label{eq:dPOdpr1}
\end{align}
The derivatives of the other parameters are obtained similarly.

\subsection{Decoding}
We use standard HMM decoder. To integrate our model with a LM, we use the WFST approach. A WFST is a finite-state machine, in which each transition has an input symbol, an output symbol and a weight. Our WFST is implemented with the FST library OpenFST \cite{Allauzen07openfst:a}, and is based on the decoding method of EESEN \cite{miao2015eesen}.

\begin{figure}
\begin{subfigure}{\linewidth}
	\centering
	\includegraphics[scale=0.32]{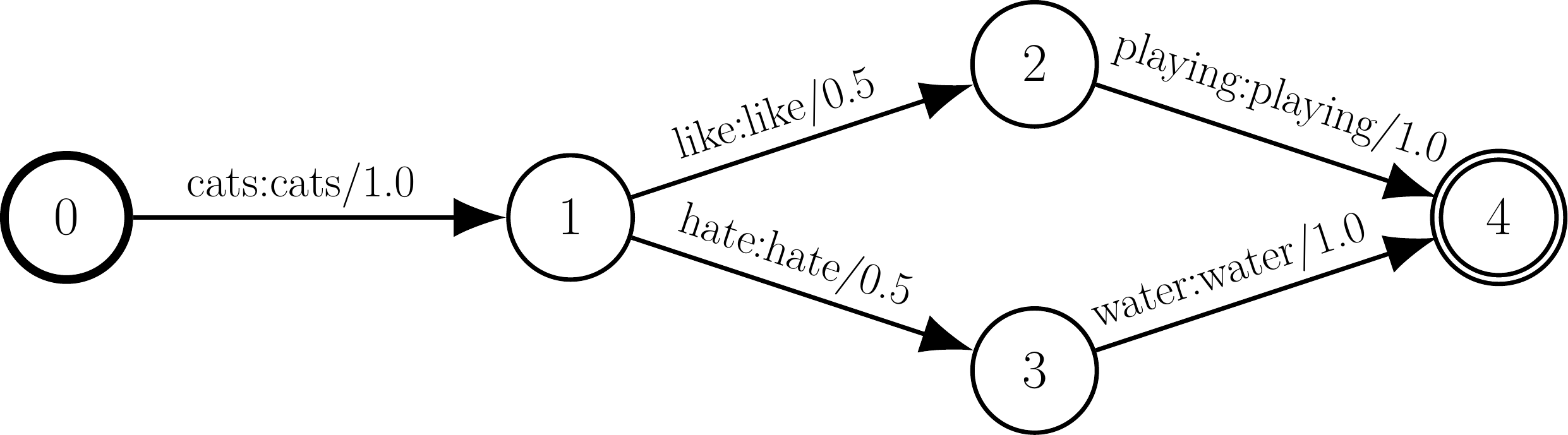}
%    \captionsetup{justification=centering}
	\caption{The $G$ WFST exemplified on the sentences: ``cats like playing'' and ``cats hate water''. The caption on each arc represents: input symbol : output symbol / weight.}
	\label{fig:G_WFST}
\end{subfigure}

\begin{subfigure}{\linewidth}
	\centering
	\includegraphics[scale=0.32]{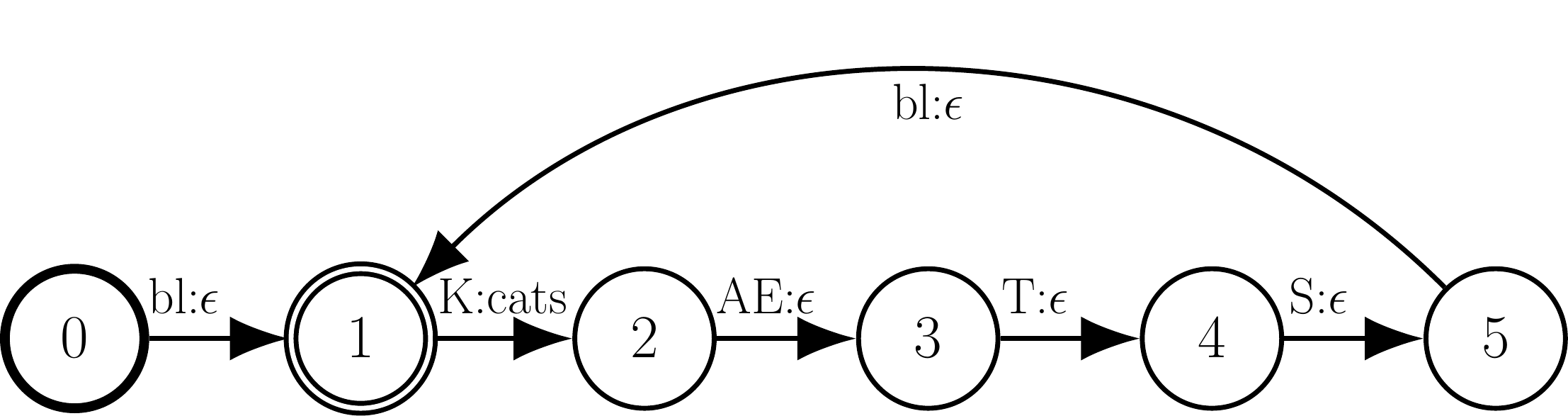}
%    \captionsetup{justification=centering}
	\caption{The $L$ WFST exemplified on the word ``cats''. The symbol $\epsilon$ represents no input/output.}
	\label{fig:L_WFST}
\end{subfigure}

\begin{subfigure}{\linewidth}
	\centering
    \hspace*{1.3cm}
	\includegraphics[scale=0.32]{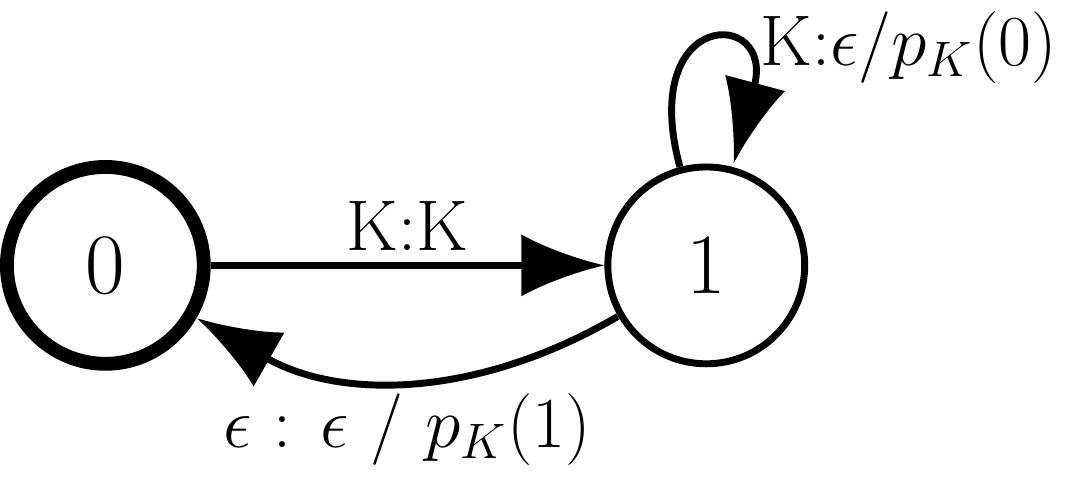}
%    \captionsetup{justification=centering}
	\caption{The $H$ WFST exemplified on the state ``K''.}
	\label{fig:H_WFST}
\end{subfigure}
\vspace{-5pt}
\caption{The $G$, $L$ and $H$ WFSTs.}
%\vspace{-15pt}
\end{figure}

We first build separate WFSTs for the LM (grammar), the lexicon and the HMM states. An example for the grammar WFST, $G$, with two possible sentences is shown in Figure \ref{fig:G_WFST}. The lexicon WFST, $L$, encodes sequences of lexicon units (in our system, phonemes) to words. It enforces the occurrence of the blank state between words, and also between identical phonemes. An example is shown in Figure \ref{fig:L_WFST}. The HMM WFST, $H$, maps a sequence of frame-level states into a single lexicon unit. It allows occurrences of repetitive states with the proper weighting according to the trained transition probabilities. An example is shown in Figure \ref{fig:H_WFST}.
Finally, we compose these graphs into a single graph, we refer to as $HLG$. The inputs to this graph are the outputs of the neural network, normalized by the priors. 
%We have thus created a decoding scheme which is consistent with our model.

\subsection{Numerical Issues}
For HMMs and CTC, the calculations of the forward and backward variables are usually performed in log-scale, as in \cite{DBLP:journals/corr/HannunCCCDEPSSCN14,miao2015eesen}. Another possibility is to use the normalization technique in \cite{rabiner1993fundamentals}. In our implementation, it was necessary to combine both approaches, with some modification of the method in \cite{rabiner1993fundamentals}, as we now describe. 
For each time frame, $t$, we first compute the forward (backward, respectively) variables according to the original recursions in log-scale, and then subtract the maximal value of the forward (backward) variables at $t$ from each forward (backward) variable.
%In the linear scale this subtraction translates to division.
A simpler variant of the above is to perform the subtraction only every certain time units.

\subsection{Voting}
As in many other disciplines, ensemble methods in ASR have been proven to be highly effective \cite{wang2017residual,deng2014ensemble,xiong2016achieving,saon2017english}. Since acoustic inference and decoding of CTC models (and ours) use so little computation, model combination is an attractive option
\cite{sak2015acoustic}. In \cite{sak2015acoustic,wang2017residual} a voting scheme is proposed for a CTC-based system using the ROVER technique \cite{fiscus1997post}. While it can be very effective \cite{wang2017residual}, the voting scheme is complicated, and requires a full decoding pass for each model, while performing the systems combination on the decoded hypothesis of each system. CTC models do not provide a good alignment between the input and output sequences, and therefore a simpler frame-level scores averaging would not yield good results. Since our method provides good alignments, we propose a simple voting scheme, based on averaging the posteriors of each model. Neither of the methods suggested in \cite{deng2014ensemble,ju2017relative} worked as well as simple averaging. Averaging the transition probabilities and the priors yields little to none gains, and therefore we only average the posteriors.

\subsection{Implementation}
We implemented our model in Tensorflow \cite{abadi2016tensorflow}. Our loss function is implemented in C++, where the samples in the mini-batch are calculated in parallel on different cores of the CPU. We have integrated the feature extraction procedures from EESEN (which uses Kaldi's scripts \cite{povey2011kaldi}) and \cite{miao2014kaldi+,TFKaldi}, the NN procedures from Tensorflow and the WFSTs decoding procedures from EESEN.
Since the computation time required by the NN is dominant, the training process with a bi-grams training LM takes only 5\%-8\% more time when using our loss function in comparison to CTC.

\subsection{Relation to Previous Work}
In \cite{povey2016purely} lattice-free training using the MMI criterion is suggested. The system uses multiple-states context dependent senones, that are obtained using decision trees. The suggested training procedure still incorporates the use of a HMM-GMM system in order to obtain alignments for a number of reasons. First, in order to split the utterances into chunks of 1.5 seconds, on which the training is applied. Second, in order to constrain the training, such that the model is only allowed to predict a state within a small window around the obtained alignment. In \cite{povey2016purely} a 4-gram train LM is used, which is trained according to the alignments. Our system operates on one-state context independent phonemes and breaks free from any use of HMM-GMM systems. In \cite{povey2016purely}, the transition probabilities are set to uniform, whereas in our system learning them provided gains in performance. Finally, we also train the prior probabilities $\omega_l$.
Out criterion is also similar to the one in \cite{collobert2016wav2letter}. However, here it is derived as a MMI estimate. Also, in \cite{collobert2016wav2letter} the values of the transition probabilities between all states have to be trained, whereas here, the probabilities $q(c,\hc)$ can be pre-trained, so that only $p_c(0)$ for all states $c$ need to be trained from the acoustics. This also allows convenient integration with higher-level LM.

\section{Experiments}
\subsection{Experimental Setup}
We conducted experiments on the Wall Street Journal (WSJ) corpus \cite{paul1992design}. The training data consists of 81 hours of transcribed speech. We use almost the same training process and NN architecture as in \cite{miao2015eesen}. Our NN architecture is a 4 layers RNN of 320 bi-directional LSTM cells \cite{hochreiter1997long} without peephole connections \cite{gers2002learning}. We use 95\% of the training set for training, and 5\% for cross validation. The inputs of the NN are 40-dimensional filterbank with delta and delta-delta coefficients. The features are normalized by mean and variance on the speaker basis. We operate on a phonemes-based system, with 72 states. The utterances in the training set are sorted by their lengths. The mini-batch size is set to 30. We use the ADAM optimizer \cite{kingma2014adam} with an initial learning rate of 0.001. We use gradient clipping \cite{pascanu2013difficulty} with a value of 50. The learning rate decay and stopping criteria are determined based on the validation WER (optimizing for each method).
%After each epoch,
%we build the WFSTs using the trained transition probabilities, and check 
%we evaluate the WER on the validation set. If the WER improves by less than 0.6, we halve %the learning rate. Training stops when the WER on the validation set deteriorates.
We test our model on the eval92 and dev93 test sets. The acoustic model scores are scaled down by a factor of 0.5-0.9, and the optimal value is chosen. All of the LMs used in training, are obtained using the training set text, and the large WSJ LM training set text.
%(we obtained improved baseline results compared to \cite{miao2015eesen} even without peephole connections).

\begin{table}[h]
%\vspace{-10pt}
%\captionsetup{aboveskip=0pt,justification=centering}
%\captionsetup{aboveskip=0pt}
\begin{center}
 \begin{tabular}{|l | c | c | c|} 
 \hline
\multicolumn{1}{|c|}{\multirow{2}{*}{Model}} & \multirow{2}{*}{LM} & \multicolumn{2}{c|}{WER\%} \\
 \cline{3-4} & & eval92 & dev93\\
 \hline\hline
 CTC, EESEN \cite{miao2015eesen} & Std. & 7.87 & 11.39 \\ 
  \hline
   CTC, ours & Std. & 7.66 & 11.61 \\
 EEMMI bi-gram & Std. & 7.37 & \textbf{10.85} \\
 EEMMI trigram & Std. & \textbf{7.05} & 11.08 \\
 \hline\hline
   Attention seq2seq \cite{chorowski2016towards} & Ext. & 6.7 & 9.7 \\  
 \hline
 CTC, ours & Ext. & 5.87 & 9.38 \\
 EEMMI bi-gram & Ext. & 5.83 & \textbf{9.02} \\
 EEMMI trigram & Ext. & \textbf{5.48} & 9.05 \\
 \hline\hline
 CTC, ROVER, 3 models \cite{wang2017residual} & Ext. & 4.29 & 7.65 \\
 \hline
EEMMI, 3 bi-grams & Ext. & 4.61 & \textbf{7.34} \\
 EEMMI, 2 bi-grams, 1 trigram & Ext. & \textbf{4.22} & 7.55 \\
 \hline
\end{tabular}
\end{center}
\vspace{-12pt}
\caption{WER of various models on the WSJ corpus. In \cite{wang2017residual} the CTC baseline is better than ours (5.48\%/9.12\% with the same architecture and ext. LM), and the eval92 set is used as a validation set.}
\label{tab:WERs}
\vspace{-18pt}
\end{table}

\subsection{Results}
Table \ref{tab:WERs} shows the results on the eval92 and dev93 sets. We consider two decoding LMs. The WSJ standard pruned trigram model (std.), and the extended-vocabulary pruned trigram model (ext.). We compare our end-to-end MMI (EEMMI) model to CTC under the same conditions. Decoding with CTC is performed using the $TLG$ WFST and prior normalization method in \cite{miao2015eesen}. Our CTC model obtains comparable results with the ones reported in \cite{miao2015eesen} \footnote{The results on the dev93 set are taken from \url{https://github.com/srvk/eesen/blob/master/asr_egs/wsj/RESULTS}}. We see a consistent improvement in WER of EEMMI using bi-grams train LM compared to CTC. It can be observed, that attention models are inferior to phonemes-based CTC and to our method. In \cite{miao2015eesen}, the hybrid HMM-DNN model achieves WERs of 7.14\%/9.91\% on the eval92/dev93 sets (std. LM), using filterbank features. In \cite{chan2015deep}, the hybrid HMM-DNN model achieves WERs of 3.47\%/6.48\% (ext. LM), using enhanced speaker adaptation. The lattice-free MMI approach \cite{povey2016purely} achieves WERs of 2.91\%/5.19\% \footnote{As reported in \url{https://github.com/kaldi-asr/kaldi/tree/master/egs/wsj/s5/local/chain}} (ext. LM), also using enhanced speaker adaptation.

\begin{figure}
\vspace{-10pt}
	\centering
	\includegraphics[width=\linewidth]{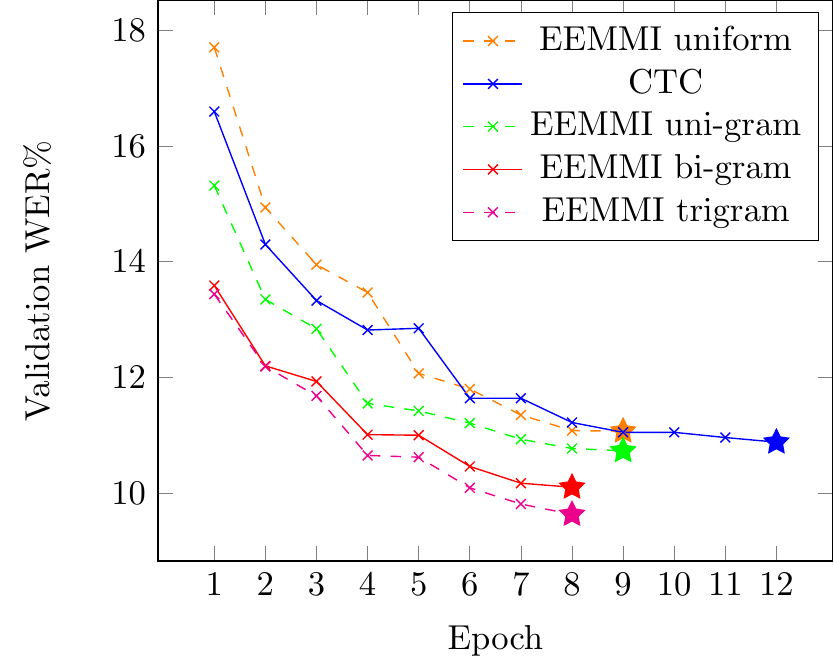}
%    \captionsetup{justification=centering}
	\vspace{-10pt}
	\caption{WER on the validation set vs. number of epochs for CTC and EEMMI.}
	\label{fig:val_WER_graph}
    %\vspace{-5pt}
\end{figure}

%We achieve a state-of-the-art result of 7.37\% WER on the eval92 test set, which is comparable with the 7.14\% WER using the Hybrd HMM-DNN model, as reported in \cite{miao2015eesen}.
Figure \ref{fig:val_WER_graph} shows the validation WER vs. epoch of CTC, and EEMMI using different train LMs: trigrams, bi-grams, uni-grams and uniform. A trigrams model achieves best results, yet the training process is more demanding than a bi-grams model (training takes approximately 3 times longer).

\subsection{Alignments and Voting}
\begin{figure}
	%\vspace{-20pt}
	\centering
	\includegraphics[width=\linewidth]{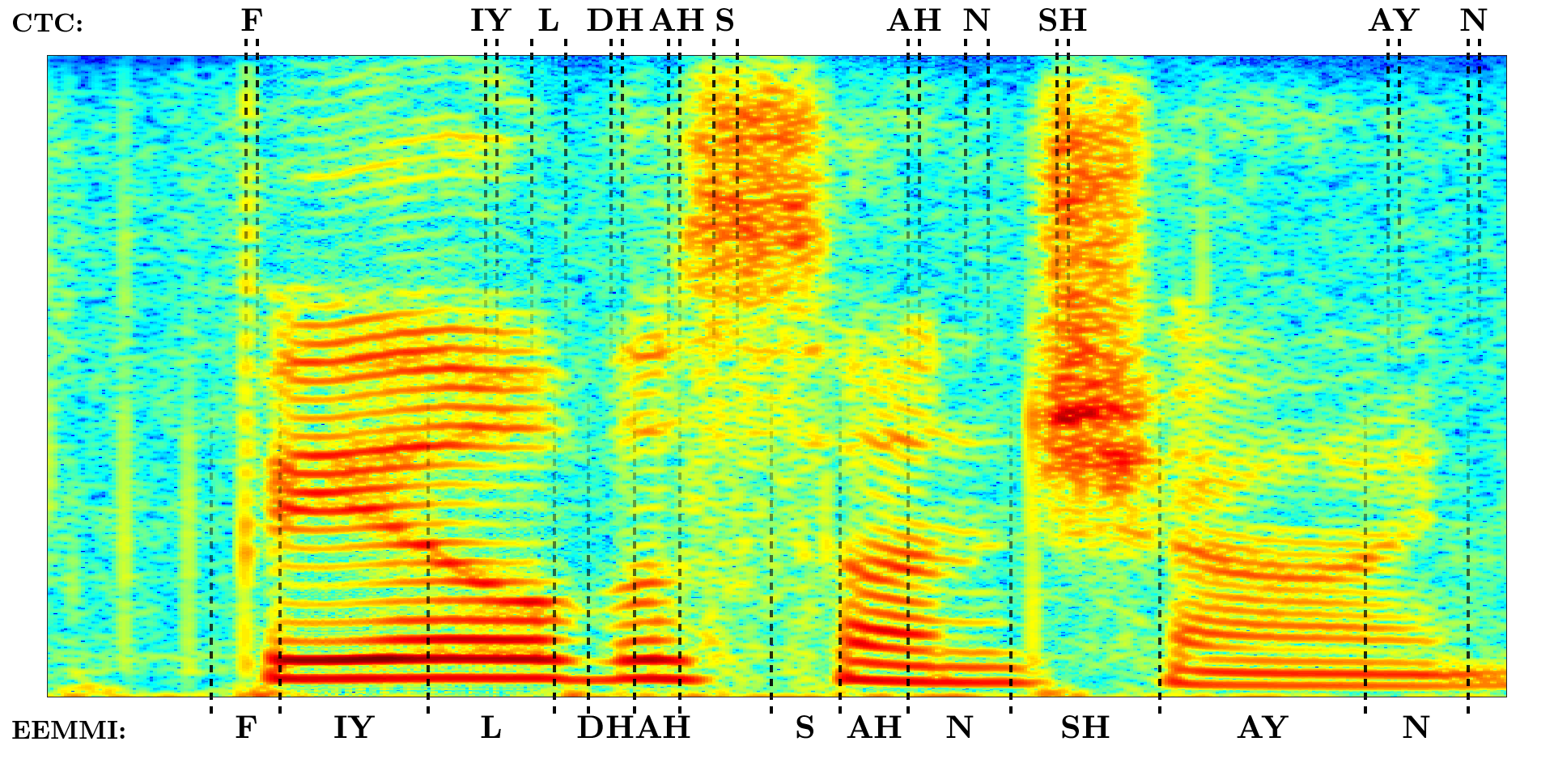}
%    \captionsetup{justification=centering}
	\vspace{-10pt}
	\caption{Alignments of CTC and our model. The sentence is ``feel the sunshine''. Blanks are not marked.}
	\label{fig:spectrogram}
    \vspace{-10pt}
\end{figure}

Figure \ref{fig:spectrogram} shows the obtained alignments on an utterance from the training set. It demonstrates a significantly improved alignment of EEMMI compared to CTC, which is due to the peaky output distributions of CTC.

These well established alignments enable the use of our simple voting scheme. We use models with the same NN architecture we described, trained separately end-to-end. Our best ensemble models, yield a considerable relative WERs reduction (WERRs) of 23\% and 18.63\% on the eval92 and dev93 test sets, resulting in WERs of 4.22\% and 7.34\% as reported in Table \ref{tab:WERs}. We note that simple posteriors averaging does not work well with CTC, even using models with identical architectures. In \cite{wang2017residual} WERRs of 19.81\% and 14.91\% were achieved, using a more complicated voting scheme, with a full decoding pass for each model. We perform decoding only once, on the averaged outputs of the models. We note that in \cite{wang2017residual} eval92 set is used as a validation set, whereas we use both eval92 and dev93 sets as test sets.

\begin{figure}[H]
%\vspace{-30pt}
\begin{subfigure}{\linewidth}
%\hspace*{-0.6cm} 
	\centering
	\includegraphics[trim={2.5cm 2cm 4cm 0},clip,scale=0.23]{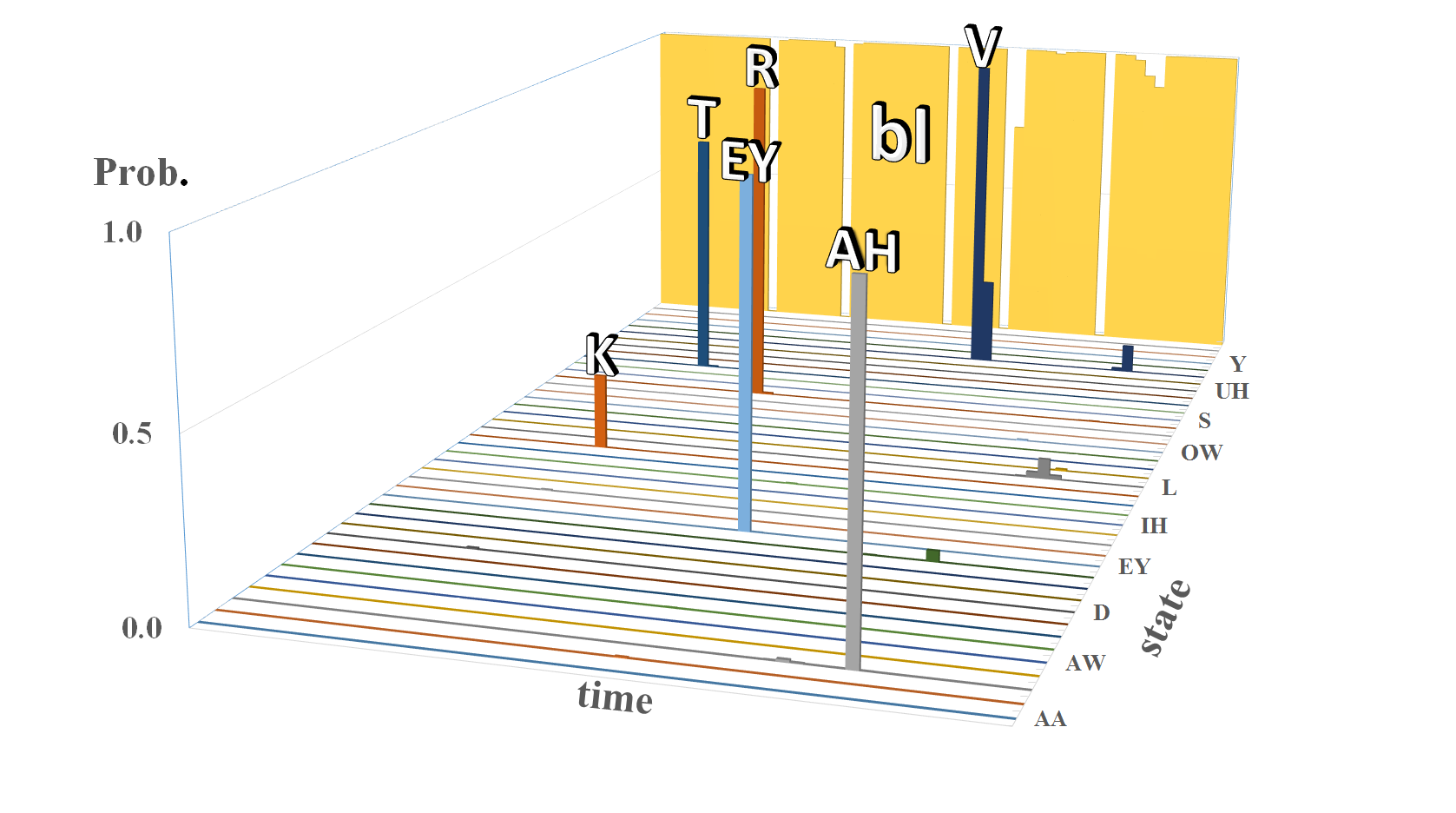}
%    \captionsetup{justification=centering}
%\vspace{-10pt}
	\caption{CTC}
	\label{fig:ctc_graph}
\end{subfigure}

\begin{subfigure}{\linewidth}
%\hspace*{-0.6cm} 
	\centering
	\includegraphics[trim={2.5cm 2cm 4cm 0},clip,scale=0.23]{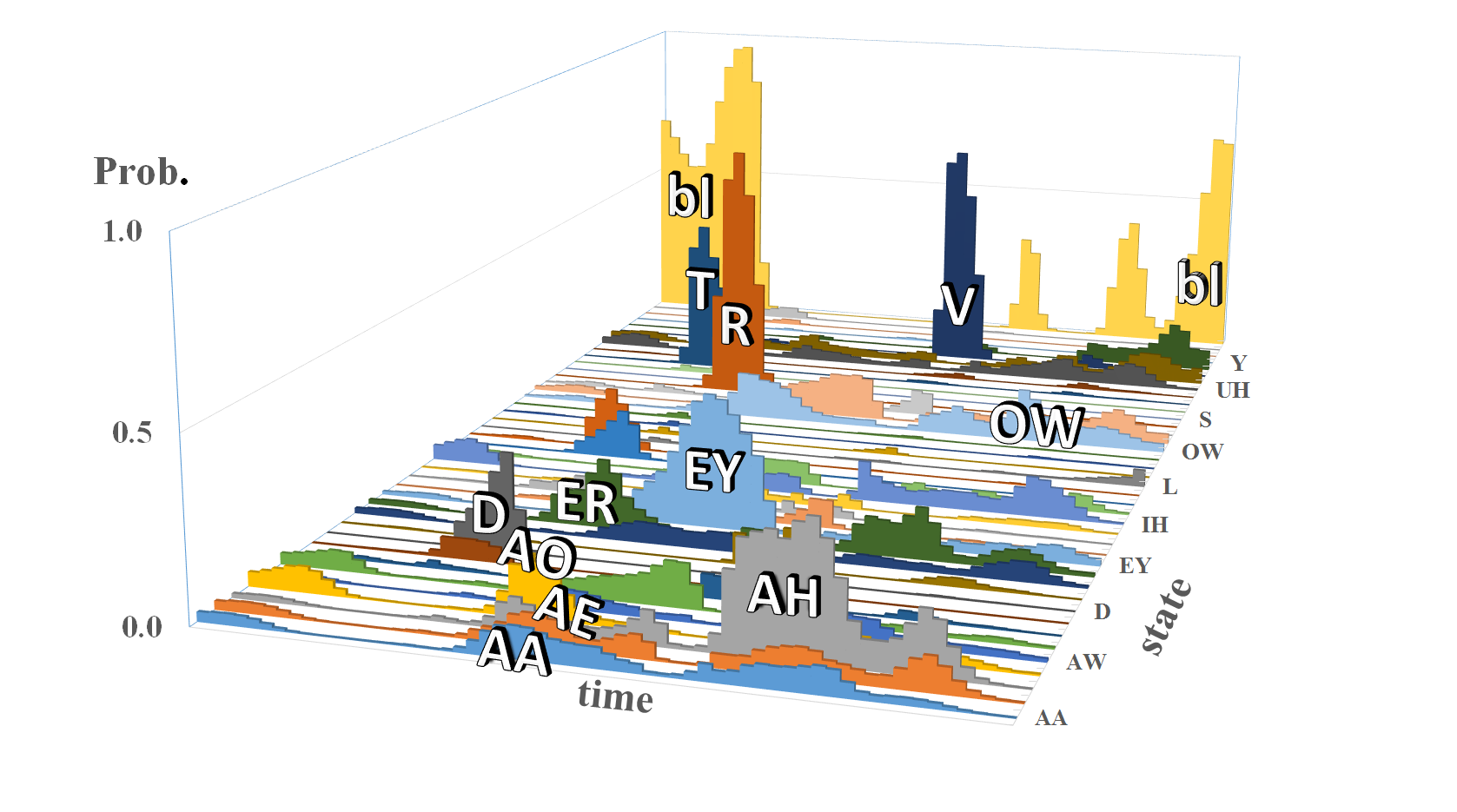}
%    \captionsetup{justification=centering}
%\vspace{-10pt}
	\caption{EEMMI}
	\label{fig:mmi_graph}
\end{subfigure}

\begin{subfigure}{\linewidth}
%\hspace*{-0.2cm}      
	\centering
	\includegraphics[trim={2.5cm 2cm 4cm 0},clip,scale=0.23]{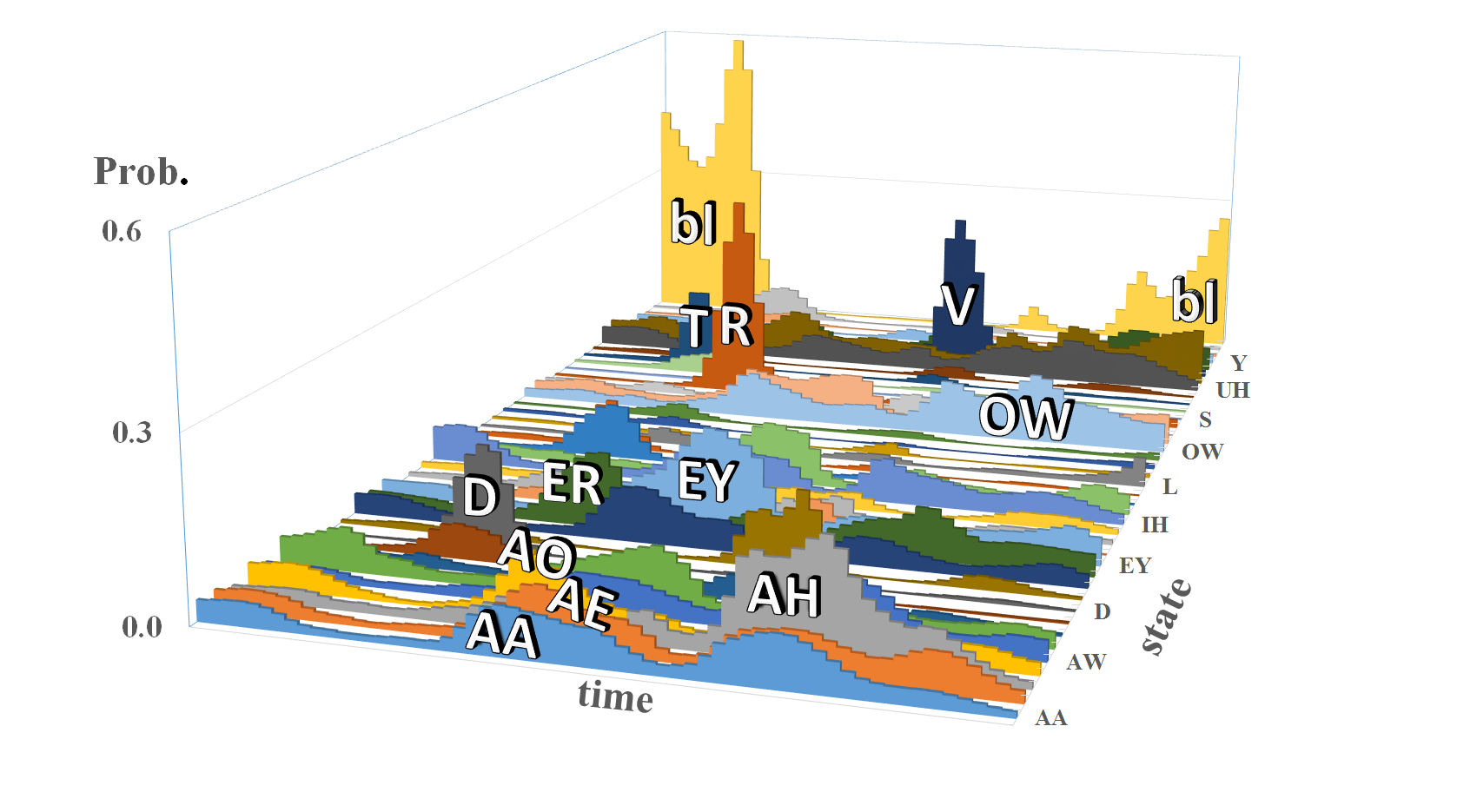}
%\vspace{-10pt}
%    \captionsetup{justification=centering}
	\caption{EEMMI ensemble}
	\label{fig:ens_graph}
\end{subfigure}
%\vspace{10pt}
\caption{Posteriors of several models corresponding to the word ``Dravo'' (``D R AE V OW'').}
\label{fig:ctc_mmi_ens}
\end{figure}

Figure \ref{fig:ctc_mmi_ens} shows the posteriors (output of the NN) of some segment from an utterance in the eval92 set. The segment corresponds to the word ``Dravo'' (``D R AE V OW''). Figure \ref{fig:ctc_graph} shows typical output distribution of CTC, where the posteriors are peaky both in time and state axes. This harms the ability of simple posteriors averaging of several models. Figure \ref{fig:mmi_graph} and \ref{fig:ens_graph} show the posteriors of a single EEMMI model and an ensemble of 3 models respectively. The ensemble method has substantially diminished the variance in the states axis of the single model.

\subsection{Decoder}
Table \ref{tab:decoder} shows a comparison between the decoders of CTC and EEMMI, with the standard LM. We demonstrate a speedup of 1.38 in decoding time. Our model's decoder graph is 43\% smaller than CTC's (due to the removal of the blank between phones), meaning that the model has a far smaller disk footprint. In \cite{miao2015eesen} it is shown that the graph size of a hybrid system operating on senones is 460 MB, and the decoding time of CTC is 3.2 times faster compared to the hybrid system. We note that the NN parameters consume 34 MB (in Tensorflow), where many different approaches may shrink this number \cite{mcgraw2016personalized}. This means that the decoder graph is the weakest link in terms of disk footprint.

\begin{table}[h]
%\vspace{-2pt}
%\captionsetup{aboveskip=0pt,justification=centering}
%\captionsetup{aboveskip=0pt}
\begin{center}
 \begin{tabular}{|c | c | c |} 
 \hline
 Model & Real-Time Factor & Graph Size (MB)\\
\hline 
  CTC &  0.0247 & 269 \\
EEMMI & 0.0179 & 154  \\
 \hline
\end{tabular}
\end{center}
\vspace{-15pt}
\caption{Decoder of CTC and EEMMI.}
\label{tab:decoder}
\vspace{-20pt}
\end{table}

\section{Conclusions}
We have presented a simplified end-to-end MMI training method for ASR. Our method aligns with the general form of CTC training, where the NN is trained to predict context independent phones. Also, our system is trained end-to-end without the need of pre-training a HMM-GMM system.
We have experimented with several simple phonemes n-grams LMs during training, and have shown that bi-grams and trigrams LMs compare favorably to CTC on the WSJ corpus. Moreover, our method presents better decoding times, and a considerable reduction of disk footprint compared to CTC and hybrid systems.
Finally, since our method provides reliable alignments, we proposed a simple voting method obtained by simply averaging the predictions of several models. This voting scheme provides a considerable reduction in WERs. Hybrid HMM-DNN speech recognizers outperform our system in terms of WER. However, due to its lower implementation and computational requirements, the proposed system may be considered in some applications.

\section{Acknowledgements}
This research was supported by the Yitzhak and Chaya Weinstein Research Institute for Signal Processing. A Tesla K40c GPU used in this research was donated by the NVIDIA Corporation.

\clearpage

\bibliographystyle{IEEEbib}
\bibliography{bibliography}
\end{document}